\documentclass[letterpaper, 10 pt, conference]{ieeeconf}
\IEEEoverridecommandlockouts                              

\usepackage[utf8]{inputenc}
\usepackage{indentfirst, times, graphicx, subfigure,
  setspace, amsmath, multirow,  epstopdf,
  amssymb, hyperref,float,caption}
\usepackage[linesnumbered]{algorithm2e}
\usepackage{accents}
\title{An MPC Walking Framework With External Contact Forces}
\author{Sean Mason$^{1}$, Nicholas Rotella$^1$, Stefan Schaal$^{1,2}$, and Ludovic Righetti$^{2,3}$
\thanks{This research was supported in part by National Science Foundation grants IIS-1205249, IIS-1017134, EECS-0926052, the European Union’s Horizon 2020 research and innovation programme (grant agreement No 780684 and European Research Council's grant No 637935), New York University, the Office of Naval Research, the Okawa Foundation, and the Max-Planck-Society. Any opinions, findings, and conclusions or recommendations expressed in this material are those of the author(s) and do not necessarily reflect the views of the funding organizations.}
\thanks{$^{1}$Computational Learning and Motor Control Lab, University of Southern California, Los Angeles, California.}
\thanks{$^{2}$Autonomous Motion Department, Max Planck Institute for Intelligent Systems, Tuebingen, Germany.}
\thanks{$^{3}$Tandon School of Engineering, New York University, New York, USA}
}

\begin{document}

\maketitle

\begin{abstract}
In this work, we present an extension to a linear Model Predictive Control (MPC) scheme that plans external contact forces for the robot when given multiple contact locations and their corresponding friction cone. To this end, we set up a two-step optimization problem. In the first optimization, we compute the Center of Mass (CoM) trajectory, foot step locations, and introduce slack variables to account for violating the imposed constraints on the Zero Moment Point (ZMP). We then use the slack variables to trigger the second optimization, in which we calculate the optimal external force that compensates for the ZMP tracking error. This optimization considers multiple contacts positions within the environment by formulating the problem as a Mixed Integer Quadratic Program (MIQP) that can be solved at a speed between 100-300 Hz. Once contact is created, the MIQP reduces to a single Quadratic Program (QP) that can be solved in real-time ($<$ 1kHz). Simulations show that the presented walking control scheme can withstand disturbances 2-3x larger with the additional force provided by a hand contact.
\end{abstract}

\section{Introduction}
In bipedal locomotion, online pattern generators have seen much success because of their ability to adapt to continually changing environment and internal states. In order to generate new plans quickly, researchers typically utilize a simplified dynamic model for fast planning of center of mass (CoM) trajectories. The plans are then tracked using a whole-body controller typically utilizing either inverse kinematics or inverse dynamics. Historically, the Linear Inverted Pendulum Model (LIPM) has been the most prominent model for biped walking because it is linear and yet effectively captures the dynamics for walking on flat surfaces. The low dimensionality and linearity of the LIPM allow the dynamics of the model to be regulated through linear control methods for compact formulations and efficient computations. 

In \cite{Kajita2003}, Kajita et. al introduced the preview control of the LIPM, which optimizes the CoM trajectory over a time-horizon. For a linear system with no inequality constraints, this is equivalent to solving the well known finite horizon Linear Quadratic Regulator (LQR), which yields a time-varying feed-forward and feedback policy. If the problem includes inequality constraints, the solution no longer provides a feedback policy that stabilize the trajectory. (One approach used to rectify this is)  Model Predictive Control (MPC). In an MPC framework, feedback is generated by continuously resolving the constrained optimization from the measured state and applying the first feed-forward command. In \cite{Wieber2006}, Center of Pressure (CoP) constraints in the form of linear inequalities are introduced and the problem is solved in a receding horizon manner to produce CoM trajectories robust to perturbations. This approach was further extended to additionally optimize over the footstep locations \cite{Herdt2010}, allowing the robot to take recovery steps and track a reference velocity. Researchers have further developed MPC methods using the LIPM to adjust footstep timing \cite{Khadiv2016,Maximo2016,Caron2016}, plan 3D trajectories \cite{Brasseur2015}, and control the Divergent Component of Motion (DCM) \cite{Englsberger2014,Hopkins2015}. A nonlinear extension also allows the robot to deal with walking around obstacles \cite{naveau17}.


\begin{figure}
\begin{center}
\includegraphics[width = 3.0in]{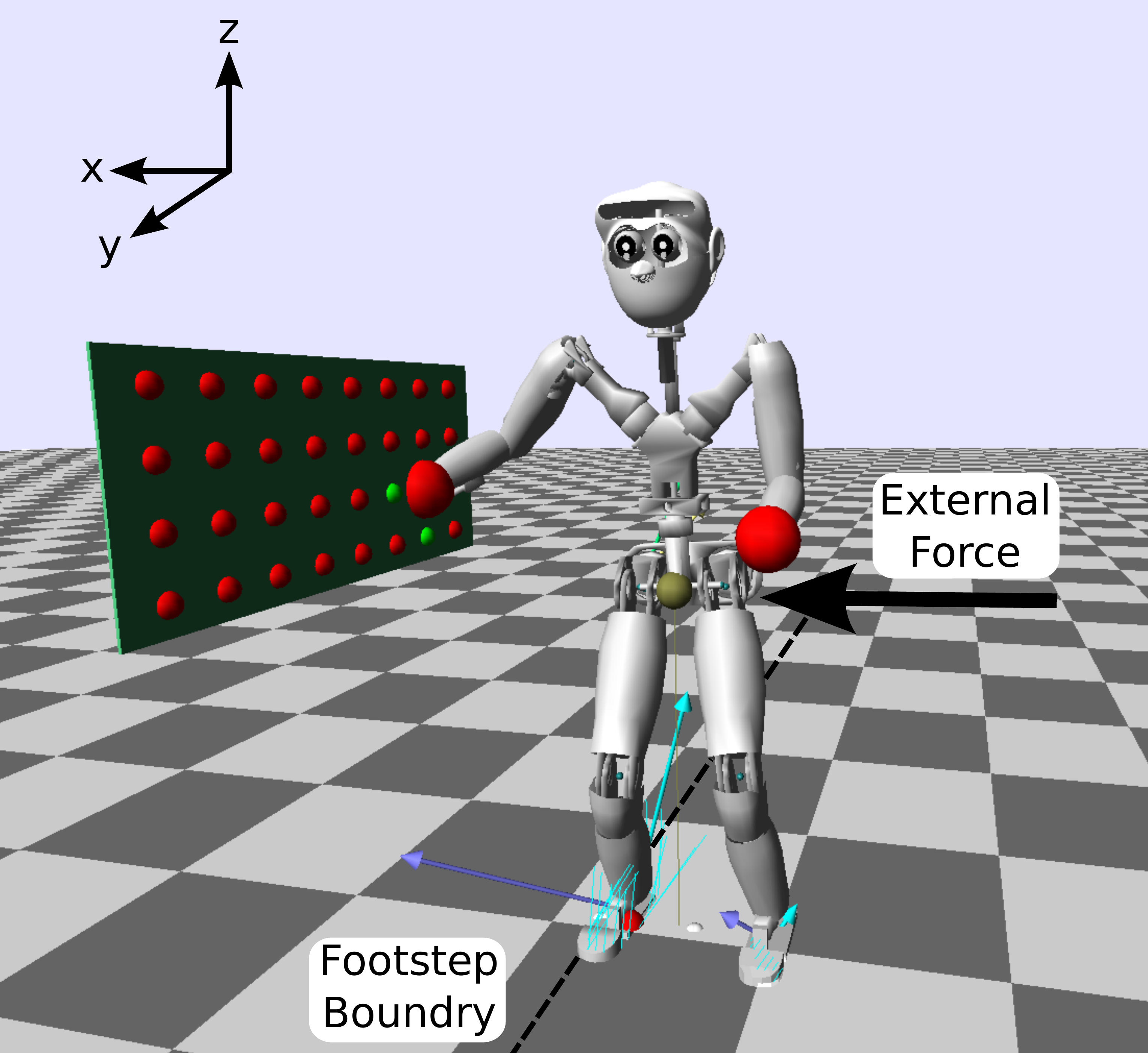}
  \caption{Visual of a push experiment setup mid push. The spheres located on the wall mark available contact points, in which green spheres specify those considered reachable.} 
  \label{fig:PushExperiment}
\end{center}
\end{figure}

The problem of designing receding horizon controllers that handle rough terrain and allow the use of hand contacts remain a largely open area. 
More general models such as the centroidal momentum dynamics have gained popularity to generate patterns allowing multi-contact behaviors \cite{JustinMomentumOptimization,Herzog-2016b} but they are not convex and are computationally more demanding. While significant progress has been made with these algorithms, they have not been applied for receding horizon control of legged robots.

Simpler models still allowing hand contacts have also been explored. In \cite{serra16}, hand contacts with the environment were introduced in an MPC problem by deriving a non-linear dynamical model of the CoM that included an external wrench. To solve the non-linear optimization problem a Newton scheme was presented in which each step of the optimization was bounded such that the intermediate solutions were always feasible. While external hand contacts were used for stability, the contact locations and timing were predetermined. Additionally, the contact considered was a grasp and rather than a push, thus neglecting friction cone constraints. In \cite{agravante16}, an external wrench is also included in the MPC scheme to design a walking pattern generator for physical collaboration. To drive cooperation with a second agent, the robot assumes the role of either a leader or follower in the task of collectively carrying an object. Contact forces are included in the dynamics and cost function such that the problem remains linear in the constraints and quadratic in the cost.

In this paper, we extend simplified walking pattern generators to exploit hand contacts when stability cannot be maintained by stepping alone. We treat the hand contact as a complement to footstep planning. To limit computational complexity and allow receding horizon control, we propose a two-part optimization that automatically calculates how to use hand contacts when the CoM and footstep adjustment alone are not sufficient. The contact locations are chosen among multiple contact points, considering both the time to reach the contact and friction cone constraints. Our decomposition maintains the computational simplicity of preview controllers based on the LIPM dynamics in the first optimization. The second part then selects a hand contact location and contact force to stabilize the robot when necessary by solving a low-dimensional mixed integer quadratic program (MIQP). The result is a controller that reactively decides when and how to use the hands of the robot to further stabilize the walking controller. Push recovery simulation experiments demonstrate that our approach can significantly improve the stability of a walking robot under strong perturbations.

\section{Newton-Euler Dynamics}
\label{sec:dynamics}
We begin by deriving the LIPM dynamics with the addition of an external contact force $f_c = [f_c^x ,f_c^y,f_c^z]$, corresponding to a hand contact located at point $p = [p^x ,p^y,p^z]$. Similar to \cite{agravante16}, we separate foot contact forces, $f_i$, from all other forces in the Newton-Euler Equations \eqref{eq:newton}, \eqref{eq:euler}.

\begin{equation}m(\ddot{c} +g) = f_c + \sum f_i \label{eq:newton}\end{equation}
\begin{equation}c \times m(\ddot{c} +g) + \dot{L} = p \times f_c + \sum s_i \times f_i \label{eq:euler}\end{equation},
where $c$ is the CoM, $L$ is the angular momentum taken about the CoM, $s_i$ is the location of the co-planar foot contacts, and the coordinate frame has the $z$ axis aligned with the gravity vector. Dividing the Euler equation by the foot contact forces in the $z$ direction, $f_i^z$, of the Newton equation and rearranging yields:

$$\frac{\sum s_i \times f_i}{\sum f_i^z}  = \frac{c \times m(\ddot{c} +g) - (p \times f_c )+\dot{L}}{m(\ddot{c}^z +g) -f_c^z}.$$




Under the LIPM assumptions, the angular momentum and the CoM height remain constant ($\dot{L} = 0$, $\ddot{c}^z = 0$), with the ground plan at zero ($s^z_i = 0$). We can then write the equations for the Zero Moment Point (ZMP) which we will refer to as $Z_\text{hand}$ (i.e. ZMP with external hand contact).

\begin{equation}
\label{eq:hand}
Z_\text{hand}^{x,y} = \frac{\sum s^{x,y}_i f^z_i}{\sum f_i^z}  =  -\frac{ m c^z \ddot{c}^{x,y} - mgc^{x,y} - p^z f^{x,y}_c + p^{x,y} f^z_c}{mg -f_c^z}
\end{equation}
Notice that if there is no contact force (i.e. $f_c = 0$) this term reduces to the familiar ZMP equations for the LIPM, $Z_{lipm}$, \cite{Kajita2003} :

\begin{equation}
\label{eq:lipm}
Z^{x,y}_\text{lipm} = \frac{\sum s^{x,y}_i f^z_i}{\sum f_i^z}  = c^{x,y} -\frac{c^z \ddot{c}^{x,y}}{g}.    
\end{equation}

\section{Shifting the ZMP Support Polygon}
To express the difference between the ZMP with and without a hand contact, we introduce the variable $\Delta Z$ such that:

\begin{equation}
    \Delta Z = Z_\text{hand} - Z_\text{lipm}
\end{equation}
or equivalently
\begin{equation}
Z_\text{hand} = Z_\text{lipm} + \Delta Z.
\end{equation}

By expressing the ZMP this way, one can see how the traditional ZMP from the LIPM shifts with the addition of an external force. As mentioned in \cite{agravante16}, we can additionally think about the implication of this force on the ZMP bounds. That is, we can rewrite

\begin{equation} \underline{Z} \leq Z_\text{lipm} + \Delta Z \leq \overline{Z} \label{eq:z_ineq} \end{equation}
as
\begin{equation}
\label{eq:shifted_zmp}
    \underline{Z}  - \Delta Z \leq Z_\text{lipm} \leq \overline{Z} -  \Delta Z.
\end{equation}

\begin{figure}
  \includegraphics[width = 3in]{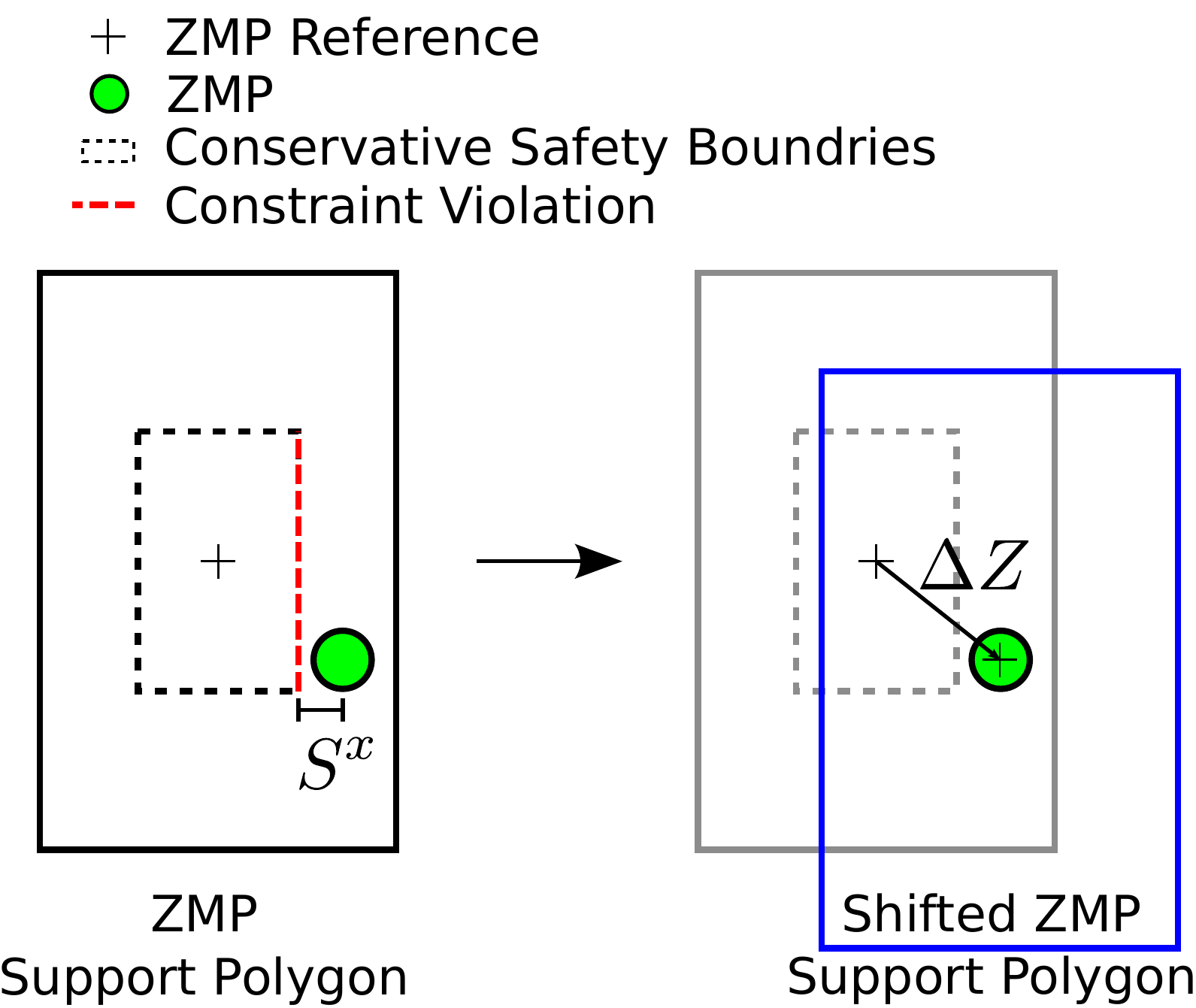}
  \caption{When the conservative ZMP bounds are violated the robot seeks to create a contact force among the available options such that the "Shifted ZMP Support Polygon", Eq. \eqref{eq:shifted_zmp} is centered at the current ZMP location.}
  \label{fig:zmp_shift}
\end{figure}

Eq. \eqref{eq:shifted_zmp} implies that it is possible to shift the effective support polygon that bounds $Z_\text{lipm}$ through the addition of an external force. Solving for $\Delta Z$ yields the nonlinear expression

\begin{equation}
\Delta Z^{x,y} = \bigg( \frac{p^z}{mg -f^z_c} \bigg) f^{x,y}_c + \bigg( \frac{-p^{x,y}  -\frac{c^z \ddot{c}^{x,y}}{g} + c^{x,y}}{mg -f^z_c} \bigg) f^z_c.
\label{eq:delta_zmp}
\end{equation}

\section{Optimization Formulation}
\label{sec:optimization}

Due to constraints on footstep locations (e.g. kinematic limits and environmental obstructions) and CoM motion (e.g. acceleration limits), it is not always feasible to keep the ZMP within the specified bounds. We remedy these situations by shifting the ZMP bounds by specifying an external force with the hand. One approach is to embed Eq. \eqref{eq:delta_zmp} into the optimization resulting in nonlinear dynamics. It also introduces the variables $f_{ci}$, $i=1, \hdots, N$, and $p$ for a total of $(3N+3)$ new decision variables, where N is the number of time-steps in the MPC preview horizon. This problem grows even further if one would like to consider points on different surfaces with different constraints (i.e. friction cone and boundary constraints). Because of the increased dimensionality, complexity, and thus the time needed to solve this larger nonlinear optimization in an online manner, we propose a more light-wight approach that remains linear by breaking the problem into two separate optimizations. In the first optimization, we decide both when an external force is necessary and what value $\Delta Z^{x,y}$ is desired to shift the support polygon. In the second optimization, we decide both the location of the hand contact $p$ and the force $f_c$. While separating the optimization into two stages will result in a sub-optimal solution, we would argue that in the context of a quick recovery motion, feasibility is more important than optimality. The result is a walking controller that hand contacts in a reactive manner when the control over the CoM and footstep locations alone is deemed insufficient. An outline of the algorithm logic is shown in Alg. 1.

\begin{algorithm}
\SetAlgoLined
 \While{Robot Has Not Fallen}{
    Find reachable contact points from list.\\
    Constrain slack variables based on time to reach contact \\
    Solve LIPM based MPC (Sec. \ref{sec:stage1}) to generate CoM, footstep, and slack trajectory. \\
  \If{$\|\text{Slack Trajectory}\| > 0$}{
    \eIf{Hand In Contact}{
    Solve QP to calculate the external force that compensates for the ZMP error (Sec. \ref{sec:stage2_force})}{
    Solve MIQP to decide the best contact location (Sec. \ref{sec:stage2_location}).\\
    Track solution with hand.}
  }
 }
 \caption{MPC Walking Scheme With Hand Contacts}
\end{algorithm}

\subsection{Stage 1: MPC of CoM Trajectory and Footstep Locations}
\label{sec:stage1}

In the first stage of the optimization, we formulate an MPC scheme to minimize CoM jerk, ZMP tracking error, and CoM velocity tracking error. The scheme is based on the well known approach from \cite{Herdt2010} so we omit the details for brevity. The only difference in our approach is that we write the ZMP constraints with the addition of a trajectory of slack variables, $S_k, k = 1 \hdots N$. We can write this constraint in the same form as Eq. \eqref{eq:shifted_zmp}, omitting the subscript $k$, as:

\begin{equation}
    \underline{Z}  - S \leq Z_\text{lipm} \leq \overline{Z} - S,
\end{equation}

where $\overline{Z}$ and $\underline{Z}$ correspond to the upper and lower bounds of a conservative support polygon (see Fig. \ref{fig:zmp_shift}). The amount we scale the support polygon boundaries is directly linked to how early the hand contact will be triggered to help. When the ZMP leaves the conservative safety boundaries, $||S|| > 0$, we treat this as an indication that additional support is necessary to maintain balance, triggering the use of a hand contact. In order to account for the fact that it will take some time for the hand to reach the contact point location, we constrain the first $N_D$ time steps of the slack variable $S$ to be zero, where $N_D$ is representative of the time-steps it will take for the hand to make contact. Furthermore, we add a weighted quadratic term to minimize $S$ in the cost function to favor solutions that only use footstep and CoM adjust for balance.

\subsection{Stage 2: Hand Contact}

Once the first optimization signifies the need for a hand contact, $||S||>0$,  a second optimization calculates the necessary contact force and location to shift the ZMP boundaries by $\Delta Z$ such that the ZMP is centered within the support polygon (see Fig. \ref{fig:zmp_shift}). While the hand has not yet created a contact, we formulate a MIQP that includes all the reachable contact point locations, along with the corresponding friction cones, to determine which, if any, is the optimal contact location to enact the shift specified by the $\Delta Z$ trajectory. Once a contact has been made, we remove all other contacts from the optimization and solve for the contact force as single QP that can be solved in real-time. We begin by formulating the problem for when the hand is already in contact.

\subsection{Determining the Contact Force}
\label{sec:stage2_force}
In Eq. \eqref{eq:delta_zmp}, we can see that with the values of $\Delta Z^{x,y}$ determined, we have two equations and three unknowns. Geometrically, the solutions for the 3D force vector form a line can be rewritten in parametric form as:

\begin{equation}
\label{eq:contact_force}
    f^{x,y}_c =  \frac{p^{x,y}  +\frac{c^z \ddot{c}^{x,y}}{g} - c^{x,y} - \Delta Z^{x,y}}{p^z} f^z_c + \frac{mg\Delta Z^{x,y}}{p^z}.
\end{equation}   
 
 Note that $f_c^{x,y}$ is a linear function of $f_c^z$ and thus the only decision variable is $f_c^z$. We include linearized friction cone constraints, by first changing the frame of our variables to be aligned with the surface normal of the contact:
 
 \begin{equation}
    \begin{bmatrix}
f^b \\    
f^t \\
f^n \end{bmatrix} = R \begin{bmatrix}
f^x \\    
f^y \\
f^z \end{bmatrix}
 \end{equation}
 
 and write the following constraints
 \begin{equation}
    \label{eq:friction_cone}
     \begin{aligned}
         f^b \leq u f^n && , &&f^b \geq -\mu f^n \\
         f^t \leq u f^n && , &&f^t \geq -\mu f^n\\
         f^n \leq f^n_\text{max} && , &&f^n \geq 0 \;\;\;\;\;\;,
     \end{aligned}
 \end{equation}

where $\mu$ is the static coefficient of friction that defines the linearized fiction cone. The cost function to optimize is then:
\begin{equation}
    \label{eq:QP1}
    \begin{aligned}
        & \underset{f^z}{\text{minimize}}
        & & (f^b)^2 + (f^t)^2 + \kappa(f^n)^2 \\
        & \text{subject to}
        & & \text{Friction Cone Constraints \eqref{eq:friction_cone}}. 
    \end{aligned}
\end{equation}

Reducing the gain $\kappa$ favors solutions that are closer to the center of the friction cone and thus more robust to slipping. This optimization formulation specified by Eq. (\ref{eq:friction_cone},\ref{eq:contact_force},\ref{eq:QP1}) will be referred to as QP1. Because the formulation of QP1 shifts the support polygon by exactly $\Delta Z$, the optimization does not consider that the ZMP does not have to be centered in the support polygon. To address this, we include the addition of the \textbf{new} slack variable $S_z = [S_z^x, S_z^y]$ that is bounded as follows:
 \begin{equation}
 \label{eq:ineq_form}
\underline{S_z} \leq S_z \leq \overline{S_z}.
 \end{equation}

Rewriting Eq. \eqref{eq:contact_force} as a function of $S_z$ and $f_z$ results in:

  \begin{equation}
 \begin{aligned} f^{x,y}_c  =  \bigg( \frac{p^{x,y}  +\frac{c^z \ddot{c}^{x,y}}{g} - c^{x,y} - \Delta Z^{x,y}}{p^z}\bigg) f^z_c \\ +\bigg( \frac{1}{p^z} \bigg) S_z^{x,y} f^z_c  + \bigg( \frac{mg}{p^z} \bigg) S_z^{x,y} + \bigg(\frac{mg \Delta Z^{x,y}}{p^z}\bigg).\end{aligned}
 \end{equation}
Unfortunately,  by introducing the new decision variable $S_z$ we have added the bilinear term $S_z f^z_c$. In order to keep the problem linear and solvable with a QP, we introduce a convex relaxation of the bilinear term by means of a McCormick envelope \cite{Castro2014}. In doing so, we create the new variables $W^{x,y} = S_z^{x,y} f^z_c$, more compactly written as $W$, and add the following new constraints to the system:
\begin{equation}
\begin{aligned}
 \underline{S_z} \, f^z_c && + &&S_z \, \underline{f^z_c} && - && \underline{S_z} \, \underline{f^z_c} && \leq && W \\
 \overline{S_z} \, f^z_c && + &&S_z \, \overline{f^z_c} && - &&  \overline{S_z} \, \overline{f^z_c} && \leq && W\\
 \overline{S_z} \, f^z_c && + &&S_z \, \underline{f^z_c} && - &&  \overline{S_z} \, \underline{f^z_c} && \geq && W\\
 S_z \, \overline{f^z_c} && + &&\underline{S_z} \, f^z_c && - &&  \underline{S_z} \, \overline{f^z_c} && \geq && W.\\
\end{aligned}
\end{equation}

\begin{figure}
\includegraphics[width = 3.4in]{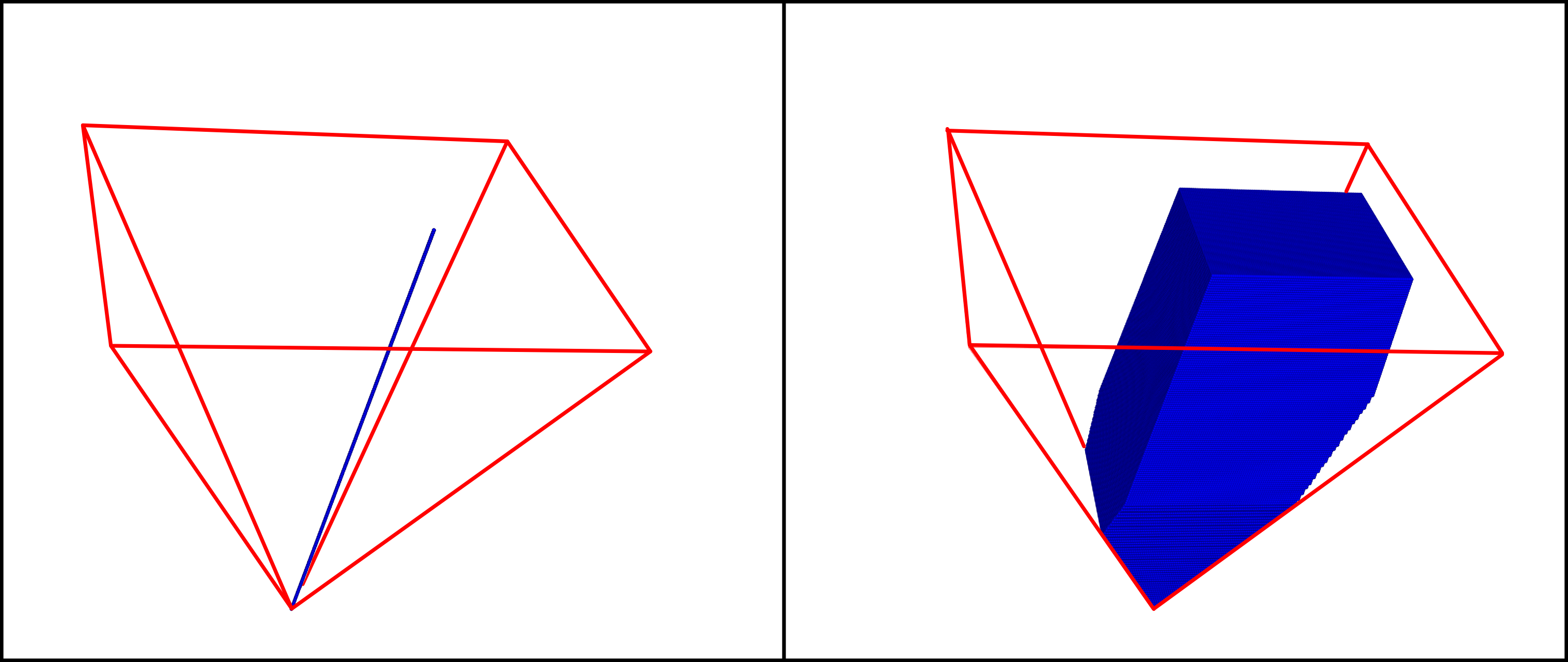}
  \caption{The left shows the solution space when we solve for the exact ZMP shift (QP1) and the right for when we formulate the optimization as QP2.}
  \label{fig:mccormick}
\end{figure}
This new formulation will be referred to as QP2.

\subsection{Determining the Contact Location}
\label{sec:stage2_location}
In order to determine the contact location, we reason about reachable contact locations and their corresponding friction cone in the form of a MIQP. To define whether a contact has been chosen, we introduce the binary variables $b_i \in {0,1}, \,\, i=1, \hdots, N_c$. We then optimize over the time steps $\hat{k}$ corresponding to the non-zero elements in the $\Delta Z$ trajectory.

\begin{equation*}
\begin{aligned}
& \underset{f^z_{i\hat{k}}, b_i}{\text{minimize}}
& & (f^b)_{i\hat{k}}^2 + (f^t)_{i\hat{k}}^2 + \kappa(f^n)_{i\hat{k}}^2 \\
& \text{subject to}
& & \text{} 
\end{aligned}
\end{equation*}

\begin{equation}
    \sum_1^{N_c} b_i = 1
\end{equation}

\begin{equation}
  b_i \Longrightarrow
  \left \{
  \begin{aligned}
    &0, && f^z_{i\hat{k}} = 0\\
    &1, &&  (\text{Friction Cone Constraints})_i
  \end{aligned} \right.
\end{equation} 

\section{Experiments}
\label{sec:Experiments}

\subsection{Optimization Benchmarks}
The experiments were conducted in this section using the optimization library Gurobi \cite{gurobi}. The average solve while searching among contacts, shown in Table \ref{tab:solve_time}, was found to vary linearly with the length of the $\Delta Z$ trajectory  and the number of contacts points. Once in contact, the optimization problem collapses down to a single QP which can be solved in 0.11 ms.

 \begin{center}
 \begin{tabular}{||c ||c |c | c| c ||} 
  \hline
    $\Delta Z$ Time-steps  & \multicolumn{4}{|c||}{Number of Contact Points ($N_c$)}   \\
 \hline
    & 5 & 10 & 20 & 40   \\
 \hline
 \hline
 16  & 5.2 ms  & 9.7 ms & 18.7 ms & 35.9 ms  \\
 \hline
 8   & 3.2 ms & 5.2 ms & 9.8 ms & 19.8 ms \\
 \hline
 4   & 1.8 ms & 2.9 ms & 5.4 ms & 10.2 ms \\
 \hline
\end{tabular}
\captionof{table}{ Solving speed for multiple contact points considering different horizon lengths.}\label{tab:solve_time}
\end{center}

\subsection{QP1 vs QP2}
As shown in Fig. \ref{fig:mccormick}, by solving for contact forces using the formulation of QP2 as opposed to QP1 we expand the feasible solution space from a line to a volume. To compare how the solutions of QP1 and QP2, we performed a parameter sweep over $\mu$ and $\Delta Z$ and evaluated 5,000 random friction cones orientations and CoM accelerations for the following ranges:  $\mu =[.1,1]$ and $\Delta Z = [-.05,.05]$, $S_z = [-.02, .02]$. In total, there were 2,000,000 test cases evaluated, resulting in the following statistics:

 \begin{center}
 \begin{tabular}{||c |c |c||} 
 \hline
  & Success Rate & Average Force (N)\\
 \hline
 Force QP1 & 36.1\% & 28.0 \\
 \hline
 Force QP2 & 73.6\% & 11.7\\
\hline
\end{tabular}
\end{center}

On average, formulation of QP2 favorably found optimal solutions for approximately twice the number test cases at approximately half the force. Because of the convex relaxation of the bilninear term used in QP2, we additionally need to check that the solutions satisfy the original problem constraints. Over the values tested, the maximum force violation was 0.037 N, which is negligible given the accuracy of force tracking capabilities of the robot seen in experiments. 

\subsection{Maximum Push Experiments}

We evaluated the proposed control scheme in simulation through a series of push recovery tests on a 31 DoF torque controlled humanoid. We are particularly interested in the relationship between the max impulse the robot can recover from with and without an external force through hand contact with the environment. Additionally, we are interested in how the number of recovery steps before hand contact affects this maximum impulse value. To test this, the robot was pushed towards a wall while walking in place from three separate distances that allow for zero, one, and two recovery steps before hand contact. Additionally, the robot footstep locations were constrained to remain at a minimum distance from the wall (0.35m) so that the upper-body would not make contact unless the arm was purposefully extended (see Fig. \ref{fig:PushExperiment}). In order to track the CoM and footstep trajectory, we use an optimization-based whole-body inverse dynamics controller formulated as a QP, similar to \cite{Feng2015,Herzog:2015a,tedrakeR}, to resolve for joint torques at 1kHz. To track the desired forces, we implemented a simple force feedback policy that was added to the resulting joint torques from the whole-body QP solver ($\tau_{qp}$): 
\begin{equation} \tau = \tau_{qp} + J_c^T (f_c + P (f_c - \hat{f_c})) \end{equation}

, where $J_c$ is the Jacobian from the CoM to the hand in contact, $\hat{f_c}$ is the measured force at the end-effector in global coordinates, and $P$ is a force-feedback gain. In the future, we plan to include these desired forces and friction cone constraints directly in the whole-body QP. The following parameters were used in these experiments: MPC horizon =  1.6 s,  MPC $\Delta t$ = 0.1 s, robot mass = 62.5 kg,  CoM height = 0.78m, single support duration = 1.0 s, double support duration = 0.1 s, $\mu$ = 1, and the support polygon boundary used in Sec. \ref{sec:stage1} scaled the foot dimensions by a factor of 0.1.

\begin{figure}[H]
\includegraphics[width = 3.4in]{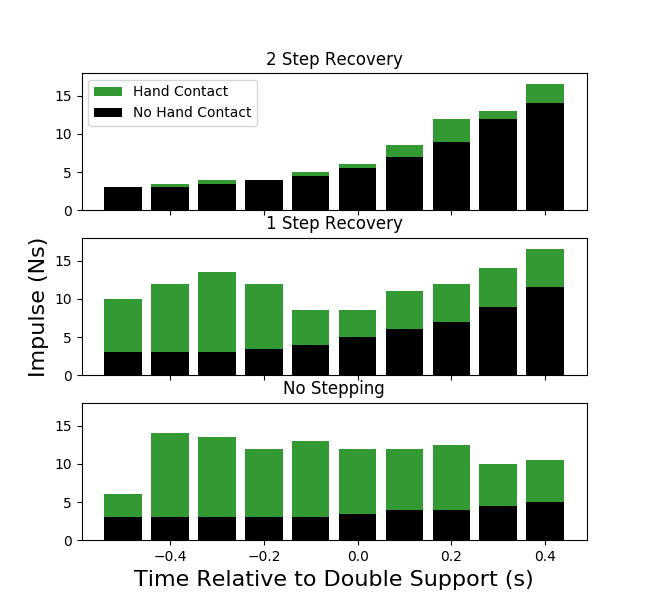}
  \caption{Maximum impulse at different times in the walking gait that the robot was able to recover from for zero, one, and two step recoveries with and without hand contacts.}
  \label{fig:push_bar_graph}
\end{figure}

 The results of the push experiments are shown in Fig. \ref{fig:push_bar_graph}. On average, using an external hand contact increased the maximum disturbance impulse by 233\%, 156\%, and 13\% for the experiments where the robot was allowed zero, one, and two recovery steps respectively. While addition of a hand contact greatly increased the walking controllers robustness to external impulses for zero and one recovery steps, in the 2-step recovery situation, the robot had a very high acceleration by the time the wall became close enough to make contact. This resulted in high desired contact forces that were difficult to track. The experiments confirm that because of the short optimization horizon and the reactive nature of this formulation, that approach presented is most effective in close proximity of contacts or when there is limited control authority through stepping. 

\begin{figure}[H]
\includegraphics[width = 3.4 in]{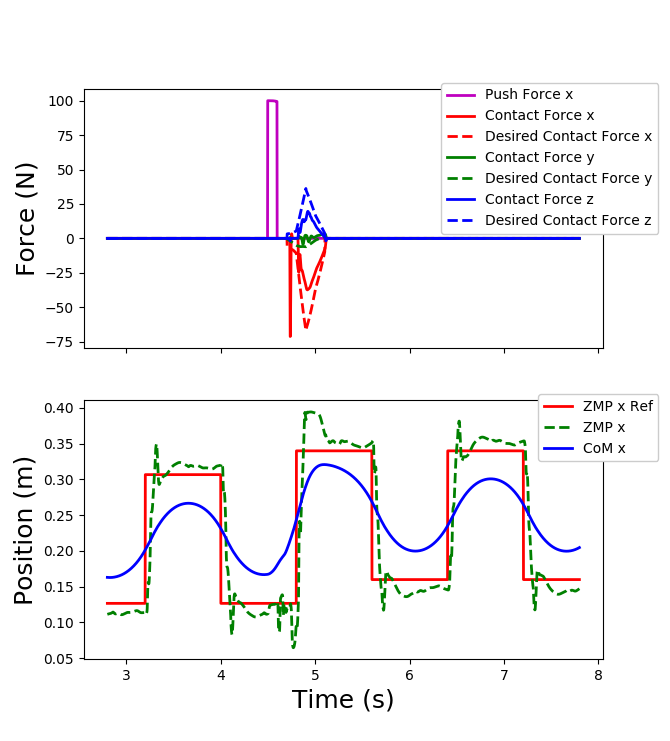}
  \caption{Top: External impulse applied to the robot (10 Ns) and the force tracking of the hand. Bottom: ZMP tracking and CoM trajectory as a result of the push. Despite imperfect force tracking, the robot is still able to recover from the impulse.}
  \label{fig:hand_force}
\end{figure}

Additionally, it is important to note that perfect force tracking was not achieved. Fig. \ref{fig:hand_force} shows the tracking of the external hand force in one of the experiments. This is important because it is very improbable that on a real humanoid perfect force tracking while making and breaking contacts with the environment will be achieved. The robustness of the approach presented comes from constantly replanning the contact force from the measured robot state. 

Further demonstrations of the behavior of this control scheme is shown in the accompanying video \footnote{\url{http://bit.ly/masonIcra2018}}. The video shows the response to pushes with and without hand contact, choosing between surfaces with different friction properties, and utilizing contacts on surfaces that are rotated at different angles.

\section{Conclusion}
In this work, we presented an extension to an MPC walking control scheme that reactively plans a desired contact location and force without any predetermined timing. In doing so, the formulation considers friction properties of the contact surfaces and the time it takes for the robot to move its hand to the contact location. To this end, we introduced a two part optimization scheme in which the first optimization plans the CoM and footsteps trajectories while the second optimization reasons about the hand contacts. The second optimization is initiated when the first optimization violates some conservative ZMP constraints. This separation into two optimization problems allowed us to reach real-time computation while maintaining low computational complexity. As shown in our experiments, there is a great benefit of using hand contact together with a walking controller when within a one recovery step proximity. Because this control scheme separates walking from hand contacts, there is the possibility to pair the second stage with different LIPM based walking pattern generators and other methods to trigger the need for hand contact. Potential directions for future research would be to use the DCM of the LIPM or incorporate an external force estimator \cite{rotella2015} to trigger the need for hand contact. 

\bibliographystyle{IEEEtran}
\bibliography{references}

\begin{thebibliography}{10}
\providecommand{\url}[1]{#1}
\csname url@samestyle\endcsname
\providecommand{\newblock}{\relax}
\providecommand{\bibinfo}[2]{#2}
\providecommand{\BIBentrySTDinterwordspacing}{\spaceskip=0pt\relax}
\providecommand{\BIBentryALTinterwordstretchfactor}{4}
\providecommand{\BIBentryALTinterwordspacing}{\spaceskip=\fontdimen2\font plus
\BIBentryALTinterwordstretchfactor\fontdimen3\font minus
  \fontdimen4\font\relax}
\providecommand{\BIBforeignlanguage}[2]{{%
\expandafter\ifx\csname l@#1\endcsname\relax
\typeout{** WARNING: IEEEtran.bst: No hyphenation pattern has been}%
\typeout{** loaded for the language `#1'. Using the pattern for}%
\typeout{** the default language instead.}%
\else
\language=\csname l@#1\endcsname
\fi
#2}}
\providecommand{\BIBdecl}{\relax}
\BIBdecl

\bibitem{Kajita2003}
S.~Kajita, F.~Kanehiro, K.~Kaneko, K.~Fujiwara, K.~Harada, K.~Yokoi, and
  H.~Hirukawa, ``Biped walking pattern generation by using preview control of
  zero-moment point,'' vol.~2, pp. 1620--1626 vol.2, Sept 2003.

\bibitem{Wieber2006}
P.~b.~Wieber, ``Trajectory free linear model predictive control for stable
  walking in the presence of strong perturbations,'' in \emph{2006 6th IEEE-RAS
  International Conference on Humanoid Robots}, Dec 2006, pp. 137--142.

\bibitem{Herdt2010}
\BIBentryALTinterwordspacing
A.~Herdt, H.~Diedam, P.-B. Wieber, D.~Dimitrov, K.~Mombaur, and M.~Diehl,
  ``{Online Walking Motion Generation with Automatic Foot Step Placement},''
  ser. Special Issue: Section Focused on Cutting Edge of Robotics in Japan
  2010, vol.~24, no. 5-6.\hskip 1em plus 0.5em minus 0.4em\relax {Taylor \&
  Francis}, 2010, pp. 719--737. [Online]. Available:
  \url{https://hal.inria.fr/inria-00391408}
\BIBentrySTDinterwordspacing

\bibitem{Khadiv2016}
M.~Khadiv, A.~Herzog, S.~A.~a. Moosavian, and L.~Righetti, ``{Step Timing
  Adjustment : A Step toward Generating Robust Gaits *},'' 2016.

\bibitem{Maximo2016}
M.~R. O.~A. Maximo, C.~H.~C. Ribeiro, and R.~J.~M. Afonso, ``{Mixed-Integer
  Programming for Automatic Walking Step Duration},'' no.~3, pp. 5399--5404,
  2016.

\bibitem{Caron2016}
S.~Caron and Q.~C. Pham, ``When to make a step? tackling the timing problem in
  multi-contact locomotion by topp-mpc,'' in \emph{2017 IEEE-RAS 17th
  International Conference on Humanoid Robotics (Humanoids)}, Nov 2017, pp.
  522--528.

\bibitem{Brasseur2015}
C.~Brasseur, A.~Sherikov, C.~Collette, D.~Dimitrov, and P.~B. Wieber, ``{A
  robust linear MPC approach to online generation of 3D biped walking
  motion},'' \emph{IEEE-RAS International Conference on Humanoid Robots}, vol.
  2015-Decem, pp. 595--601, 2015.

\bibitem{Englsberger2014}
J.~Englsberger, T.~Koolen, S.~Bertrand, J.~Pratt, C.~Ott, and
  A.~Albu-Schäffer, ``Trajectory generation for continuous leg forces during
  double support and heel-to-toe shift based on divergent component of
  motion,'' pp. 4022--4029, Sept 2014.

\bibitem{Hopkins2015}
M.~A. Hopkins, D.~W. Hong, and A.~Leonessa, ``{Humanoid locomotion on uneven
  terrain using the time-varying divergent component of motion},''
  \emph{IEEE-RAS International Conference on Humanoid Robots}, vol. 2015-Febru,
  pp. 266--272, 2015.

\bibitem{naveau17}
M.~Naveau, M.~Kudruss, O.~Stasse, C.~Kirches, K.~Mombaur, and P.~Souères, ``A
  reactive walking pattern generator based on nonlinear model predictive
  control,'' \emph{IEEE Robotics and Automation Letters}, vol.~2, no.~1, pp.
  10--17, Jan 2017.

\bibitem{JustinMomentumOptimization}
J.~Carpentier, S.~Tonneau, M.~Naveau, O.~Stasse, and N.~Mansard, ``A versatile
  and efficient pattern generator for generalized legged locomotion,'' in
  \emph{2016 IEEE International Conference on Robotics and Automation (ICRA)},
  May 2016, pp. 3555--3561.

\bibitem{Herzog-2016b}
A.~Herzog, S.~Schaal, and L.~Righetti, ``Structured contact force optimization
  for kino-dynamic motion generation,'' in \emph{2016 IEEE/RSJ International
  Conference on Intelligent Robots and Systems (IROS)}, Oct 2016, pp.
  2703--2710.

\bibitem{serra16}
D.~Serra, C.~Brasseur, A.~Sherikov, D.~Dimitrov, and P.~B. Wieber, ``A newton
  method with always feasible iterates for nonlinear model predictive control
  of walking in a multi-contact situation,'' in \emph{2016 IEEE-RAS 16th
  International Conference on Humanoid Robots (Humanoids)}, Nov 2016, pp.
  932--937.

\bibitem{agravante16}
D.~J. Agravante, A.~Sherikov, P.~B. Wieber, A.~Cherubini, and A.~Kheddar,
  ``Walking pattern generators designed for physical collaboration,'' in
  \emph{2016 IEEE International Conference on Robotics and Automation (ICRA)},
  May 2016, pp. 1573--1578.

\bibitem{Castro2014}
P.~Castro and F.~de~Cincias, ``Tightening piecewise mccormick relaxations
  through partition-dependent bounds for non-partitioned variables,'' 11 2014.

\bibitem{gurobi}
\BIBentryALTinterwordspacing
I.~Gurobi~Optimization, ``Gurobi optimizer reference manual,'' 2016. [Online].
  Available: \url{http://www.gurobi.com}
\BIBentrySTDinterwordspacing

\bibitem{Feng2015}
S.~Feng, E.~Whitman, X.~Xinjilefu, and C.~G. Atkeson, ``{Optimization based
  full body control for the atlas robot},'' \emph{IEEE-RAS International
  Conference on Humanoid Robots}, vol. 2015-February, no.~Id, pp. 120--127,
  2015.

\bibitem{Herzog:2015a}
A.~Herzog, N.~Rotella, S.~Mason, F.~Grimminger, S.~Schaal, and L.~Righetti,
  ``Momentum control with hierarchical inverse dynamics on a torque-controlled
  humanoid,'' \emph{Autonomous Robots}, vol.~40, no.~3, pp. 473--491, Mar.
  2016.

\bibitem{tedrakeR}
R.~Tedrake \emph{et~al.}, ``A summary of team mit's approach to the virtual
  robotics challenge,'' in \emph{Robotics and Automation (ICRA), 2014 IEEE
  International Conference on}, May 2014, pp. 2087--2087.

\bibitem{rotella2015}
N.~Rotella, A.~Herzog, S.~Schaal, and L.~Righetti, ``Humanoid momentum
  estimation using sensed contact wrenches,'' in \emph{2015 IEEE-RAS 15th
  International Conference on Humanoid Robots (Humanoids)}, Nov 2015, pp.
  556--563.

\end{thebibliography}

\end{document}